\documentclass[letterpaper, 10 pt, conference]{ieeeconf}  

\IEEEoverridecommandlockouts                              

\usepackage{amsmath} 
\usepackage{txfonts} 

\usepackage[linesnumbered,vlined,ruled,boxed,commentsnumbered]{algorithm2e}
  \usepackage{algpseudocode}
  \usepackage{bm} 

\usepackage{threeparttable}
\usepackage{multirow}
\usepackage{bbm}

\usepackage{color, colortbl}

\usepackage{verbatim}

\newcommand{\fref}[1]{Figure \ref{#1}}

\usepackage{cleveref}

\usepackage{multirow,booktabs,color,soul,threeparttable}
\definecolor{hl}{rgb}{0.75,0.75,0.75}
\sethlcolor{hl}

\usepackage{array}

\usepackage{cite}

\usepackage{graphicx}

\usepackage{scalefnt}

\title{\LARGE \bf
An Inverse Modeling Constrained Multi-Objective Evolutionary Algorithm Based on Decomposition}

\author{Lucas R. C. Farias$^{1}$ $^{2}$ $^{3}$ and Aluizio F. R. Ara\'ujo$^{1}$
\thanks{$^{1}$Centro de Inform\'atica, Universidade Federal de Pernambuco, Recife, Brazil,
        {\tt\small \{lrcf,aluizioa\}@cin.ufpe.br}}%
\thanks{$^{2}$Universidade Católica de Pernmabuco, Recife, Brazil,
        {\tt\small lucas.farias@unicap.br}}%
\thanks{$^{3}$Centro de Estudos e Sistemas Avançados do Recife, Recife, Brazil,
        {\tt\small lrcf@cesar.org.br}}%
}

\begin{document}

\maketitle
\thispagestyle{empty}
\pagestyle{empty}

\begin{abstract}

This paper introduces the inverse modeling constrained multi-objective evolutionary algorithm based on decomposition (IM-C-MOEA/D) for addressing constrained real-world optimization problems. Our research builds upon the advancements made in evolutionary computing-based inverse modeling, and it strategically bridges the gaps in applying inverse models based on decomposition to problem domains with constraints. The proposed approach is experimentally evaluated on diverse real-world problems (RWMOP1-35), showing superior performance to state-of-the-art constrained multi-objective evolutionary algorithms (CMOEAs). The experimental results highlight the robustness of the algorithm and its applicability in real-world constrained optimization scenarios.

\end{abstract}


\section{Introduction}

A multi-objective optimization problem (MOP) is characterized by a set of Pareto optimal solutions, representing a compromise between the different objectives. Such solutions form the Pareto set (PS) in the decision space with the correspondent Pareto front (PF) in the objective space. Various optimization problems in real-world applications hold a number of conflicting objectives and multiple complex constraints, such as the robot gripper optimization problem, water resources management, gearbox design, and process synthesis problem \cite{kumar2021benchmark}. This kind of problem is denoted as constrained MOP (CMOP). 

CMOPs bring more challenges than their unconstrained counterparts due to the coexistence of multiple objectives and constraints. The CMOP is defined as \cite{liang2022survey}:
\begin{equation}
\label{eq:moop}
\begin{array}{ll}
\min & \mathbf{f}(\mathbf{x})=\left(f_1(\mathbf{x})  \text{ } f_2(\mathbf{x})  \ldots  f_m(\mathbf{x})\right)^T \\
\text { s.t. } & \left\{\begin{array}{l}    
g_j(\mathbf{x}) \leq 0, j=1, \ldots, l \\
h_j(\mathbf{x})=0, j=l+1, \ldots, k \\
\mathbf{x}=\left(x_1 \text{ } x_2 \ldots x_d\right)^T \in \Omega
\end{array}\right.
\end{array}
\end{equation}
where $\mathbf{x}$ is a $d$-dimensional real-valued decision vector in the decision space $\varmathbb{R}^d$, and $\mathbf{f}(\mathbf{x})$ is an $m$-dimensional objective vector in the objective space $\varmathbb{R}^m$; $\Omega \subset \varmathbb{R}^n$ defines the feasible region of the decision space; $g_j(\mathbf{x})$ $\leq$ 0 is the $j$-th inequality constraint; $h_j(\mathbf{x})$ $=$ 0 is the ($j-l$)-th equality constraint; and $l$ and ($k-l$) denote the number of inequality and equality constraints.  

In recent decades, multi-objective evolutionary algorithms (MOEAs) have emerged as a popular approach for generating a set of non-dominated solutions in a single run to solve MOPs \cite{liang2022survey}. Existing MOEAs can be classified according to their selection methods: Pareto-based, decomposition-based, and indicator-based approaches \cite{cheng2015multiobjective, liang2022survey, shen2023inverse}. These MOEAs are effective solvers for unconstrained multi-objective optimization problems. However, when applied to CMOPs, they need a constraint handling technique (CHT), a selection mechanism that can handle constraints. CMOPs are more prevalent in practical application scenarios than unconstrained MOPs. Thus, considerable effort has been devoted to designing CHTs and specific mechanisms to promote the emergence and development of various CMOEAs \cite{yuan2021indicator}.

While MOEAs often require population diversity to store non-dominated solutions within the population or in an external archive, studies with model-based MOEAs aim to loosen such a diversity requirement. Model-based approaches focus on constructing a probabilistic model in the decision space throughout the evolutionary process \cite{cheng2015multiobjective}. Despite their effectiveness, the model-based MOEAs still rely on maintaining a candidate solution set, such as an external archive, to preserve non-dominated solutions obtained during the search. Another strategy to relax the diversity necessity involves constructing a regression model to identify solutions enhancing the diversity in the final generation \cite{cheng2015multiobjective, farias2021moea}.
    
The just mentioned alternatives inspired Cheng et al. \cite{cheng2015multiobjective} to introduce an MOEA using Gaussian process-based inverse modeling (IM-MOEA). It stands out as a Pareto-based evolutionary algorithm based on an inverse model that maps the objective space into the decision space. This approach is promising when the decision-making process is essential for solving MOPs. The inverse mapping is an optimization process that becomes an optimal allocation of solutions within the Pareto front, simplifying the identification of suitable solutions. The optimal allocation refers to an approximation set of the PF that provides maximum information about the underlying PF, and it releases adherence to a uniform distribution of Pareto-optimal solutions \cite{farias2021moea}.

Subsequent research has expanded upon the foundation laid by IM-MOEA \cite{lin2018dynamic,gholamnezhad2020inverse, zhang2020enhanced, farias2021moea, shen2023inverse, da2023learning, da2023framework}. These works propose improvements such as adaptive weight vectors in population clustering \cite{lin2018dynamic,zhang2020enhanced}, E-IM-MOEA utilizes an external population and a biased reproduction operator \cite{lin2018dynamic}, IM-MOEA-RF replaces the Gaussian process with Random Forest \cite{gholamnezhad2020inverse}, and recently some works have extended inverse models based on Pareto dominance to deal with CMOP \cite{shen2023inverse, da2023learning, da2023framework}. IM-MOEA/D proposes a decomposition-based approach tested as competitive on large-scale many-objective problems \cite{farias2021moea}. Decomposition-based inverse modeling to deal with CMOPs has yet to be proposed to the best of our knowledge.

Bearing in mind the potential of inverse modeling for solving MOPs, we think it can do well for CMOPs since it acts directly in the objective space, allowing a better distribution mapping of the constraint behavior than in decision space. In this context, we propose the inverse modeling constrained multi-objective evolutionary algorithm based on decomposition (IM-C-MOEA/D) that incorporates a constraint handling technique that adjusts its search strategy based on the presence and severity of constraints, actively guiding the optimization process to enable the solutions' feasibility and improving convergence toward the Pareto front \cite{shen2023inverse, jain2013evolutionary}. To validate our approach, we conducted experiments using a set of real-world CMOPS, namely RWMOP1-RWMOP35 \cite{kumar2021benchmark}, in which the number of objectives ranges from 2 to 5, the number of variables varies between 2 and 30, and has up to 29 constraints. Our experimental results suggest the superior performance of the proposed method compared to six state-of-the-art MOEAs: paired offspring generation-based constrained evolutionary algorithm (POCEA) \cite{he2020paired}, large-scale multi-objective competitive swarm optimization algorithm (LMOCSO) \cite{tian2019efficient}, reference vector guided evolutionary algorithm (RVEA) \cite{cheng2016reference}, many-objective evolutionary algorithm based on dominance and decomposition (MOEA/DD) \cite{li2014evolutionary}, IM-MOEA \cite{cheng2015multiobjective}, and IM-C-MOEA, an extended version of IM-MOEA to deal with constraints that we also proposed.

The fundamental concept is elaborated in Section 2, followed by a detailed description of the primary components of the IM-C-MOEA/D. Section 3 presents the experimental results of the proposed algorithm in comparison to six other CMOEAs on 35 real-world instances. Finally, Section 4 provides the conclusion of the study and outlines future work.



\section{IM-C-MOEA/D}
\label{sec:proposed-method}

\begin{figure}
        \centering
        \includegraphics[width=0.75\linewidth]{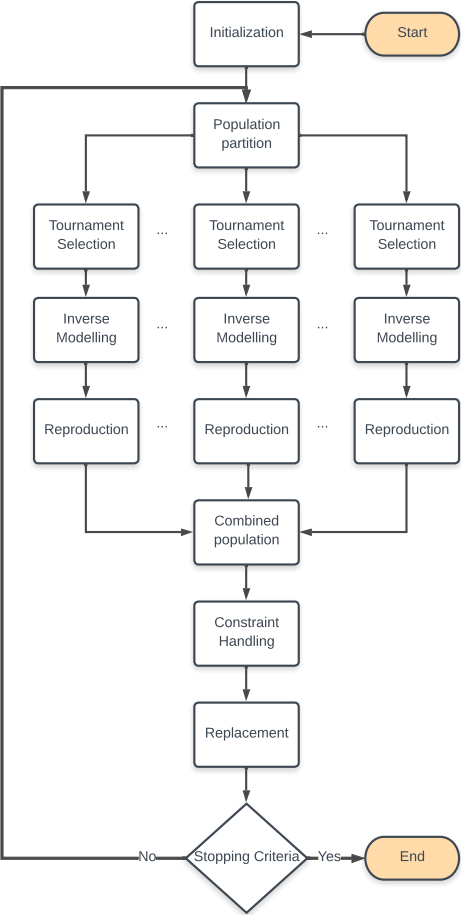}
        \caption{Flowchart of execution of the IM-C-MOEA/D.}
        \label{fig:flux-IMCMOEAD}
\end{figure}

This section discusses the details of the inverse modeling constrained multi-objective evolutionary algorithm based on decomposition (IM-C-MOEA/D). A high-level scheme is initially shown in \fref{fig:flux-IMCMOEAD}. The main items that compose it are the following:

\begin{itemize}
    \item Initialization: As most decomposition-based MOEAs do, we initialize the individuals of the population, $P$, weight vectors, $\boldsymbol{\lambda}$, and the reference point, $\boldsymbol{z}$. In particular, the weight vectors are initially generated using the Das and Dennis sampling (Subsection II.A).
    
    \item Population partition: The subpopulations are created by the $k$-means algorithm, followed by a tournament selection of solutions in the subpopulation (Subsection II.B).
    
    \item Inverse model: Each subpopulation has an inverse model, a mapping from the objective space into the decision space (Subsection II.C).

    \item Constraint handling: The constraint handling procedure to compare viable and unviable solutions for replacement purposes is described in Subsection II.D.
    
    \item Decomposition-based global replacement: The survivor selection is based on decomposition using a global replacement, which associates each new solution with the most suitable weight vector (Subsection II.E).
    
    \item Stopping Criteria: The maximum number of generations or fitness evaluations.
\end{itemize}
The details of the IM-C-MOEA/D components and their computational complexity (Subsection II.F) are below.

\subsection{Generating Weights}

The weight vectors are generated uniformly distributed over a unit simplex using the method proposed by Das and Dennis. This approach is present in most variants of decomposition-based MOEAs \cite{wang2014replacement}. The number of weight vectors to be generated, $N$, is calculated as $N = \binom{H + m - 1}{m - 1}$, where $H$ is the number of divisions of each axis and $m$ is the number of objectives. The value of $H$ cannot be lower than $m$ to avoid creating intermediate points. Moreover, generating weight vectors may become computationally costly for more than three objectives \cite{deb2014evolutionary}.

\subsection{Population Partition}

IM-MOEA/D builds the inverse models (IMs) in the reproduction stage; firstly, the $k$-means algorithm creates clusters according to the position in the objective space \cite{farias2021moea}. Then, IM-C-MOEA/D applies a binary tournament to select solutions, with priority to feasible ones, in all clusters \cite{jain2013evolutionary}. There is an IM for each group.

\subsection{Inverse Modeling}
\label{subsec:inverseModeling}

Conventional model-based algorithms aim to estimate the distribution of the candidate solutions in the decision space. In contrast, inverse model-based algorithms are built to represent the inverse mapping from the objective space to the decision space. After constructing the IMs, evenly distributed candidate solutions can be directly sampled from the objective space and mapped into the decision space \cite{cheng2015multiobjective}. However, estimating the inverse mapping from the $m$-dimensional objective space to the $d$-dimensional decision space can be challenging. Therefore, a multivariate IM is decomposed into several univariate regression models:

\begin{equation}
\label{eq:inverseModeling}
P(X \vert Y) \approx \prod^{d}_{i=1}{(P(x_i \vert f_j) + \epsilon_{j,i})},
\end{equation}
where $j = 1, 2, ..., m$, $i = 1, 2, ..., d$, $P(x_i \vert f_j)$ is an univariate model that represents the inverse mapping from objective function $f_j$ to decision variable $x_i$, and $\epsilon _{j, i}$ is an error term. We consider that $\epsilon _{j, i}$ $\sim$ $\mathcal{N} (0,(\sigma _n)^2 )$ is an additive Gaussian noise. Consequently, each univariate model, together with the error term achieved using the Gaussian process, is a way for modeling both the global regularity and the local randomness in the distribution of the candidate solutions during the search \cite{cheng2015multiobjective}.

\subsection{Constraint Handling Approach}

IM-C-MOEA/D extends IM-MOEA/D \cite{farias2021moea} incorporating a constrained handling approach suggested by Jain and Deb \cite{jain2013evolutionary, shen2023inverse}. This modification enables the proposed IM-C-MOEA/D to handle CMOPs effectively. Therefore, in IM-C-MOEA/D, when a new candidate solution $\mathbf{y}$ (an offspring) is compared with a randomly selected neighbor $\mathbf{x}$, the constraint violations (CV) of both solutions are verified before applying solution substitution based on the Tchebycheff metric. Depending on the feasibility of solutions, the following four scenarios can occur:

\begin{enumerate}
\item If $\mathbf{y}$ is infeasible and $\mathbf{x}$ is feasible, then $\mathbf{x}$ remains unchanged;
\item If $\mathbf{y}$ is feasible and $\mathbf{x}$ is infeasible, then $\mathbf{x}$ is replaced by $\mathbf{y}$;
\item If both $\mathbf{y}$ and $\mathbf{x}$ are infeasible, then $\mathbf{x}$ is replaced by $\mathbf{y}$ if  the constraint violation of the latter is lower than that of the former;
\item If both solutions are feasible, $\mathbf{x}$ is replaced by $\mathbf{y}$ when $\mathbf{y}$ has a better Tchebycheff metric value than $\mathbf{x}$.
\end{enumerate}

Such modifications prioritize feasible and small-CV solutions in the population without inserting any new parameter to the algorithm \cite{jain2013evolutionary}.

\begin{algorithm}
\caption{IM-C-MOEA/D}
\label{alg:framework}

Initialize the population $P$ and a weight vectors set $\lambda$;

Define $T$ = 0.1$N$;

Determine the $T$ neighbors in $B(i)$ for each $\boldsymbol{\lambda}_i \in \lambda$;

Calculate the reference point $\mathbf{z}$ based on the population $P$;

\For{each $Gen = 1$ ... $Gen_{max}$}
{
    Partition the population $P(Gen)$ using $k$-means to create subparent populations $P_1$($Gen$), ..., $P_K$($Gen$);
    
    $O \gets \emptyset$;
    
    \For{$each$ $k \in \{1,...,K\}$}
    {
        Apply the tournament selection operator to $P_k(Gen)$

        Build the inverse model through the Gaussian process using subpopulation $P_k(Gen)$;

        Sample candidate solutions $O_k(Gen)$ in the objective space and map them back to the decision space using the inverse models; 
        
        Perform mutation on $O_k(Gen)$;
        
        $O \gets O \cup O_k(Gen)$;
    }
    
    $\mathbf{z} \gets min (\mathbf{z},\mathbf{f}(O(Gen)))$;
    
    \For{$each$ $\mathbf{o} \in O(Gen)$}
    {
        Find the most appropriate weight vector $\boldsymbol{\lambda_i}$ for $\mathbf{o}$ using Equation \eqref{eq:globalReplacement};

        Calculate the constraint violation ($CV$) of $\mathbf{o}$ and each solution in $B(i)$;
    
        \For{$each$ $j \in B(i)$}
        {
            \If {$g^{TCH}(\mathbf{o} \vert \boldsymbol{\lambda}_j, \mathbf{z}) \leq g^{TCH}(P_j \vert \boldsymbol{\lambda}_j, \mathbf{z})$ and $CV(\mathbf{o}) \leq CV(P_j)$}
            {
            	$P_j \gets \mathbf{o}$;
            }
        }
    }
}
\textbf{return} $P$;
\end{algorithm}

\subsection{Decomposition-Based Replacement}

The Tchebycheff (TCH) decomposition is defined as:
\begin{equation}
\begin{split}
\label{eq:tchebycheff}
\text{minimize } & g^{TCH}(\mathbf{x} \vert \boldsymbol{\lambda} , \mathbf{z}) = \max _{1 \leq j \leq m} (\lambda_j \vert f_j(\mathbf{x}) - z_j \vert),\\
&\text{subject to } \mathbf{x} \in \Omega
\end{split}
\end{equation}
where $m$ is the number of objectives and $\mathbf{z}$ is the utopian reference point, i.e. $z_j$ = min $\{ f_j(\mathbf{x})\vert \mathbf{x} \in \Omega \}$, for every $j = 1,...,m$. The $m$-dimensional weight vector is defined as $\boldsymbol{\lambda}$ = ($\lambda_1$ ... $\lambda_m$)$^T$, $\sum_{1}^{m} \lambda_j = 1$ and $\lambda_j \geq 0$, for all $j \in 1,...,m$. Different Pareto-optimal solutions can be obtained by altering weight vectors using the TCH approach \cite{wang2014replacement}.

After generating a new solution $\mathbf{o}$, the global replacement scheme determines the most suitable weight vector $i$ using the TCH decomposition:
\begin{equation}
    \label{eq:globalReplacement}
    i=\arg \min _{1 \leq k \leq N} \left\{g^{TCH}\left(\mathbf{o} \mid \lambda_{k}, \mathbf{z}\right)\right\}.
\end{equation}
The algorithm then verifies the neighborhood of the $\boldsymbol{\lambda}_i$, searching for solutions that can be improved by the offspring $\mathbf{o}$. If $g^{TCH} (\mathbf{o} \mid \lambda_{j}, \mathbf{z})$ is better than $g^{TCH}  (\mathbf{x}_j \mid \lambda_{j}, \mathbf{z})$, the current solution $\mathbf{x}_j$ in the $T$-neighborhood of the weight vector  $\boldsymbol{\lambda}_j$ is replaced by $\mathbf{o}$ \cite{wang2014replacement}.

\subsection{Algorithm Framework}
\label{sec:alg-framework}

The pseudo-code for IM-C-MOEA/D is illustrated in Algorithm \ref{alg:framework}. IM-C-MOEA/D initializes the population and weight vectors, determines the neighborhood structure for each weight vector, and calculates the ideal reference point (lines 1-4). The population is partitioned into $K$ groups using $k$-means (line 6), and new individuals are generated for each partition (lines 7-13). The global replacement is applied to identify the most suitable weight vector, and the individuals in the neighborhood of the most suitable weight vector are examined using the constraint handling approach and decomposition function (lines 14-20). The algorithm stops when the maximum number of generations is reached. The maximum computational complexity of IM-C-MOEA/D is determined by the time cost of population partitioning using $k$-means ($\mathcal{O}(N^2)$), reproduction by inverse modeling ($\mathcal{O}(N^3)$), and global replacement ($\mathcal{O} (mN^2)$). The overall computational complexity is $\mathcal{O}(N^3)$.

\begin{table*}[ht]
\scalefont{0.87}
\centering
\setlength{\tabcolsep}{3pt} 
\caption{Mean and standard deviation of the Hypervolume on RWMOP1-35. Best performances are Highlighted. Settings for the population size ($N$), the number of objectives ($m$), decision variables ($d$), the maximum of fitness evaluation ($FE$), and quantity of inequality ($ng$) and equality ($nh$) constraints.}
\label{tab:imc}
\begin{tabular}{@{}lcccccccccc@{}}
\hline
\multicolumn{1}{c}{Problem}                                      & $N$ & $m$ & $d$ & $ng$ & $nh$ & $FE$  & IM-C-MOEA                                     & \multicolumn{1}{c|}{IM-MOEA}                             & IM-C-MOEA/D                                   & IM-MOEA/D           \\ \hline
\multicolumn{11}{c}{Mechanical Design Problems}                                                                                                                                                                                                                                           \\
Pressure Vessel Design                                           & 80  & 2   & 4   & 2    & 2    & 20000 & \hl{4.6051e-1 3.13e-2 +}         & \multicolumn{1}{c|}{0.0000e+0 (0.00e+0)}                 & \hl{5.8560e-1 7.39e-3 +}        & 0.0000e+0 (0.00e+0) \\
Vibrating Platform Design                                        & 80  & 2   & 5   & 5    & 0    & 20000 & \hl{3.6766e-1 7.79e-3 +}        & \multicolumn{1}{c|}{0.0000e+0 (0.00e+0)}                 & \hl{3.3757e-1 3.56e-2 +}        & 0.0000e+0 (0.00e+0) \\
Two Bar Truss Design                                             & 80  & 2   & 3   & 3    & 0    & 20000 & \hl{8.6356e-1 6.18e-3 +}        & \multicolumn{1}{c|}{1.9427e-1 (2.74e-1)}                 & \hl{7.1513e-1 4.62e-2 +}        & 0.0000e+0 (0.00e+0) \\
Welded Beam Design                                               & 80  & 2   & 4   & 4    & 0    & 20000 & \hl{7.8057e-1 2.14e-2 +}        & \multicolumn{1}{c|}{2.9166e-1 (2.62e-1)}                 & \hl{5.0765e-1 9.09e-2 +}        & 0.0000e+0 (0.00e+0) \\
Disc Brake Design                                                & 80  & 2   & 4   & 4    & 0    & 20000 & \hl{4.2608e-1 7.55e-4 +}        & \multicolumn{1}{c|}{2.6687e-1 (1.78e-2)}                 & \hl{4.2795e-1 8.56e-4 +}        & 2.3416e-1 (6.83e-3) \\
Speed Reducer Design                                             & 80  & 2   & 7   & 11   & 0    & 20000 & \hl{2.7533e-1 2.31e-4 +}        & \multicolumn{1}{c|}{0.0000e+0 (0.00e+0)}                 & \hl{2.7468e-1 3.49e-4 +}        & 0.0000e+0 (0.00e+0) \\
Gear Train Design                                                & 80  & 2   & 4   & 1    & 0    & 20000 & \hl{4.7110e-1 1.89e-3 +}        & \multicolumn{1}{c|}{4.6305e-1 (2.97e-3)}                 & \hl{4.7906e-1 2.08e-3 +}        & 4.7157e-1 (5.36e-3) \\
Car Side Impact Design                                           & 105 & 3   & 7   & 9    & 0    & 26250 & \hl{2.3507e-2 3.31e-4 +}        & \multicolumn{1}{c|}{2.0817e-2 (8.62e-4)}                 & \hl{2.3871e-2 4.29e-4 +}        & 1.6289e-2 (1.46e-3) \\
Four Bar Plane Truss                                             & 80  & 2   & 4   & 0    & 0    & 20000 & \hl{3.6739e-1 8.50e-3 $\approx$} & \multicolumn{1}{c|}{3.6587e-1 (7.52e-3)}                 & \hl{4.0321e-1 2.05e-3 +}        & 3.3149e-1 (1.72e-2) \\
Two Bar Plane Truss                                              & 80  & 2   & 2   & 2    & 0    & 20000 & 8.3546e-1 (4.69e-3) $\approx$   & \multicolumn{1}{c|}{\hl{8.3677e-1 3.71e-3}} & \hl{8.4480e-1 5.81e-4 +}        & 8.3696e-1 (2.18e-3) \\
Water Resources Management                                       & 212 & 5   & 3   & 7    & 0    & 53000 & \hl{8.4717e-2 2.04e-3 +}        & \multicolumn{1}{c|}{7.1942e-2 (3.29e-3)}                 & \hl{9.0599e-2 1.47e-3 +}        & 7.5898e-2 (2.86e-3) \\
Simply Supported I-beam Design                                   & 80  & 2   & 4   & 1    & 0    & 20000 & \hl{4.9200e-1 1.73e-2 +}        & \multicolumn{1}{c|}{2.0577e-1 (1.20e-1)}                 & \hl{4.0147e-1 5.92e-2 +}        & 2.1557e-4 (1.18e-3) \\
Gear Box Design                                                  & 105 & 3   & 7   & 11   & 0    & 26250 & \hl{8.6440e-2 2.76e-4 +}        & \multicolumn{1}{c|}{1.1527e-3 (6.31e-3)}                 & \hl{8.7439e-2 1.99e-4 +}        & 0.0000e+0 (0.00e+0) \\
Multiple Disk Clutch Brake Design                                & 80  & 2   & 5   & 8    & 0    & 20000 & \hl{5.3033e-1 1.75e-2 +}        & \multicolumn{1}{c|}{1.3503e-1 (1.15e-1)}                 & \hl{4.9617e-1 3.39e-2 +}        & 4.4961e-2 (3.41e-2) \\
Spring Design                                                    & 80  & 2   & 3   & 8    & 0    & 20000 & \hl{4.9915e-1 6.04e-3 +}        & \multicolumn{1}{c|}{0.0000e+0 (0.00e+0)}                 & \hl{2.8940e-1 7.85e-2 +}        & 0.0000e+0 (0.00e+0) \\
Cantilever Beam Design                                           & 80  & 2   & 2   & 2    & 0    & 20000 & \hl{7.1653e-1 1.01e-2 +}        & \multicolumn{1}{c|}{6.1425e-1 (7.77e-2)}                 & \hl{5.6697e-1 7.51e-2 +}        & 1.1608e-1 (3.79e-3) \\
Bulk Carrier Design                                              & 105 & 3   & 6   & 9    & 0    & 26250 & \hl{1.9777e-1 2.98e-2 +}        & \multicolumn{1}{c|}{1.6645e-2 (6.33e-2)}                 & \hl{1.9138e-1 4.20e-2 +}        & 1.9777e-2 (5.91e-2) \\
Front Rail Design                                                & 80  & 2   & 3   & 3    & 0    & 20000 & \hl{3.8177e-2 3.28e-4 +}        & \multicolumn{1}{c|}{3.6332e-2 (8.30e-4)}                 & \hl{4.0302e-2 8.05e-5 +}        & 3.9935e-2 (2.12e-4) \\
Multi-product Batch Plant                                        & 105 & 3   & 10  & 10   & 0    & 26250 & \hl{3.2244e-1 6.73e-3 +}        & \multicolumn{1}{c|}{0.0000e+0 (0.00e+0)}                 & \hl{3.2377e-1 4.52e-3 +}        & 0.0000e+0 (0.00e+0) \\
Hydro-static Thrust Bearing Design                               & 80  & 2   & 4   & 7    & 0    & 20000 & 0.0000e+0 (0.00e+0) $\approx$   & \multicolumn{1}{c|}{0.0000e+0 (0.00e+0)}                 & 0.0000e+0 (0.00e+0) $\approx$   & 0.0000e+0 (0.00e+0) \\
Crash Energy Management for High-speed Train                     & 80  & 2   & 6   & 4    & 0    & 20000 & \hl{3.1130e-2 2.39e-4 +}        & \multicolumn{1}{c|}{3.1048e-2 (1.61e-4)}                 & \hl{3.1405e-2 6.63e-5 +}        & 3.0760e-2 (7.36e-6) \\
\multicolumn{11}{c}{Chemical Engineering Problems}                                                                                                                                                                                                                                        \\
Haverly’s Pooling Problem                                        & 80  & 2   & 9   & 2    & 4    & 20000 & 0.0000e+0 (0.00e+0) $\approx$   & \multicolumn{1}{c|}{0.0000e+0 (0.00e+0)}                 & 0.0000e+0 (0.00e+0) $\approx$   & 0.0000e+0 (0.00e+0) \\
Reactor Network Design                                           & 80  & 2   & 6   & 1    & 4    & 20000 & \hl{1.1034e-1 1.59e-1 +}        & \multicolumn{1}{c|}{0.0000e+0 (0.00e+0)}                 & 0.0000e+0 (0.00e+0) $\approx$   & 0.0000e+0 (0.00e+0) \\
Heat Exchanger Network Design                                    & 105 & 3   & 9   & 0    & 6    & 26250 & 0.0000e+0 (0.00e+0) $\approx$   & \multicolumn{1}{c|}{0.0000e+0 (0.00e+0)}                 & 0.0000e+0 (0.00e+0) $\approx$   & 0.0000e+0 (0.00e+0) \\
\multicolumn{11}{c}{Process, Design, and Synthesis Problems}                                                                                                                                                                                                                              \\
Process Synthesis Problem                                        & 80  & 2   & 2   & 2    & 0    & 20000 & \hl{2.1984e-1 2.41e-3 +}        & \multicolumn{1}{c|}{1.7608e-1 (1.07e-1)}                 & \hl{2.3916e-1 1.36e-3 +}        & 2.3282e-1 (1.85e-5) \\
Process Synthesis and Design Problem                             & 80  & 2   & 3   & 1    & 1    & 20000 & \hl{1.6593e-1 2.45e-2 +}        & \multicolumn{1}{c|}{0.0000e+0 (0.00e+0)}                 & \hl{1.7561e-1 1.94e-2 +}        & 0.0000e+0 (0.00e+0) \\
Process Flow Sheeting Problem                                    & 80  & 2   & 3   & 3    & 0    & 20000 & \hl{1.9524e+2 2.44e+2 +}        & \multicolumn{1}{c|}{1.0780e-1 (3.31e-1)}                 & \hl{1.4589e+5 4.60e+5 +}        & 5.2170e+2 (2.30e+3) \\
Two Reactor Problem                                              & 80  & 2   & 7   & 4    & 4    & 20000 & 0.0000e+0 (0.00e+0) $\approx$   & \multicolumn{1}{c|}{0.0000e+0 (0.00e+0)}                 & 0.0000e+0 (0.00e+0) $\approx$   & 0.0000e+0 (0.00e+0) \\
Process Synthesis Problem                                        & 80  & 2   & 7   & 9    & 0    & 20000 & \hl{6.9427e-1 1.18e-1 +}        & \multicolumn{1}{c|}{0.0000e+0 (0.00e+0)}                 & \hl{7.5814e-1 3.19e-2 +}        & 0.0000e+0 (0.00e+0) \\
\multicolumn{11}{c}{Power Electronics Problems}                                                                                                                                                                                                                                           \\
Synch. Optimal PWM of 3-level Inverters  & 80  & 2   & 25  & 24   & 0    & 80000 & \hl{3.4240e-2 1.07e-1 +}        & \multicolumn{1}{c|}{0.0000e+0 (0.00e+0)}                 & \hl{1.0268e-1 2.36e-1 +}        & 0.0000e+0 (0.00e+0) \\
Synch. Optimal PWM of 5-level Inverters  & 80  & 2   & 25  & 24   & 0    & 80000 & \hl{1.5029e-2 5.78e-2 $\approx$} & \multicolumn{1}{c|}{0.0000e+0 (0.00e+0)}                 & \hl{2.0911e-2 1.15e-1 $\approx$} & 0.0000e+0 (0.00e+0) \\
Synch. Optimal PWM of 7-level Inverters  & 80  & 2   & 25  & 24   & 0    & 80000 & \hl{5.5420e-2 1.53e-1 +}        & \multicolumn{1}{c|}{0.0000e+0 (0.00e+0)}                 & \hl{9.9896e-2 2.59e-1 +}        & 0.0000e+0 (0.00e+0) \\
Synch. Optimal PWM of 9-level Inverters  & 80  & 2   & 30  & 29   & 0    & 80000 & 0.0000e+0 (0.00e+0) $\approx$   & \multicolumn{1}{c|}{0.0000e+0 (0.00e+0)}                 & 0.0000e+0 (0.00e+0) $\approx$   & 0.0000e+0 (0.00e+0) \\
Synch. Optimal PWM of 11-level Inverters & 80  & 2   & 30  & 29   & 0    & 80000 & 0.0000e+0 (0.00e+0) $\approx$   & \multicolumn{1}{c|}{0.0000e+0 (0.00e+0)}                 & 0.0000e+0 (0.00e+0) $\approx$   & 0.0000e+0 (0.00e+0) \\
Synch. Optimal PWM of 13-level Inverters & 80  & 2   & 30  & 29   & 0    & 80000 & \hl{2.7882e-2 7.35e-2 +}        & \multicolumn{1}{c|}{0.0000e+0 (0.00e+0)}                 & 0.0000e+0 (0.00e+0) $\approx$   & 0.0000e+0 (0.00e+0) \\ \hline
                                                                 &     &     &     &      &      &       & 26/0/9                                        & \multicolumn{1}{c|}{}                                    & 26/0/9                                        &                     \\ \hline
\end{tabular}
\end{table*}
 
\section{Validation of the Algorithm}
\label{sec:validation}
	
We introduce the benchmark problems used for the validation experiments and the parameter settings used in all CMOEAs. Then, we present the performance metric for assessing the quality of the approximate Pareto-optimal solutions generated by each competing algorithm. At last, the experimental results are described and analyzed.

\subsection{Experimental Setup}
\label{sec:experimentalSetup}

The validation process was designed to evaluate the performance of the IM-C-MOEA/D on a range of real-world problems, RWMOP1 to RWMOP35 \cite{kumar2021benchmark}.  The features are detailed in columns 2 to 7 of Table I. In particular, the definition of the population size and number of function evaluations in each problem was extracted from the original RWMOP paper \cite{kumar2021benchmark}. The problems are categorized into Mechanical Design, Chemical Engineering, Process, Design and Synthesis, and Power Electronics.

IM-C-MOEA/D and IM-C-MOEA were implemented in the PlatEMO, a MATLAB-based platform in which the other CMOEAs codes are available. The experiments run in a computer with 32 gigabytes of RAM and a 3.60GHz 8-core Intel Core i9-9900KF processor. 

We used the parameters informed by the literature for the CMOEAs. The variation operators for MOEA/DD \cite{li2014evolutionary} and RVEA \cite{cheng2016reference} are SBX and polynomial mutation \cite{jain2013evolutionary}, with both distribution indices set to 20 and the probabilities of crossover and mutation set to $1.0$ and $1/d$ ($d$ is the number of decision variables). The penalty parameter $\alpha$ of APD in LMOCSO \cite{tian2019efficient} is set to 2. In POCEA \cite{he2020paired}, the parameter $K$ is set to 5. For the inverse models, IM-MOEA \cite{cheng2015multiobjective}, IM-MOEA/D \cite{farias2021moea}, IM-C-MOEA, and IM-C-MOEA/D, the two main parameters are the number of groups $K$ set to 10, and the model group size $L$, set to 3 or 2 when the problem has only two decision variables.

The performance of each algorithm is evaluated through the widely-used metric, namely the hypervolume (HV) \cite{zitzler1999multiobjective}.

\textbf{HV}: Given a reference point $\mathbf{z}$ = ($z_1$ ...  $z_n$)$^T$ dominated by all Pareto-optimal solutions, the HV of a set of solutions $P$ is defined as the volume of the objective space dominated by all solutions in $P$, bounded by $\mathbf{z}$:
    \begin{equation}
    \label{eq:hypervolume}
    HV(P, \mathbf{z}) = Vol \left( \bigcup_{\mathbf{x} \in P} [\mathbf{f}(\mathbf{x}), \mathbf{z}] \right),
    \end{equation}
where $Vol(\cdot)$ denotes the Lebesgue measure, and [$\mathbf{f}$($\mathbf{x}$), $\mathbf{z}$] represents the hyperrectangle formed by the fitness of the solution $\mathbf{x}$ and the reference point $\mathbf{z}$. The larger the HV, the better the approximation quality of $P$. We use a reference point 10\% higher than the upper bound of the PF for our experiments. To reduce the computational complexity of determining HV if $m > 4$, we use the Monte Carlo method with 1,000,000 sampling points to approximate its value.

The tests were run 30 times independently, and each performance mean and standard deviation were recorded. The Wilcoxon rank-sum test was conducted to analyze the results with a significance level 0.05. The findings were interpreted with symbols +, -, and $\approx$, meaning that the result by another MOEA is significantly better, worse, or statistically equivalent to the results obtained by IM-C-MOEA/D.

\subsection{Experiments}

Tables I and II present the experimental results of IM-C-MOEA/D compared with the chosen CMOEAs for solving the RWMOP1-35 \cite{kumar2021benchmark}. For each test problem, the number of objectives ($m$) and decision variables ($d$) vary from 2 to 5 and 2 to 30. The maximum number of constraints is 29.

Table I presents the HV values achieved by the original models (IM-MOEA and IM-MOEA/D) and their versions for handling constraints: IM-C-MOEA/D and IM-C-MOEA share features regarding the constraint violation criterion as Jain and Deb \cite{shen2023inverse, jain2013evolutionary} have proposed. The experimental results show that MOEAs with constraint treatments performed better in 26 out of 35 problems. The performance was statistically equivalent for the remaining nine instances. 

Table II presents experimental results achieved by the IM-C-MOEA/D and five state-of-the-art CMOEAs used for solving real-world multi-objective problems, RWMOP1-35. Out of the six algorithms, IM-C-MOEA/D achieved the best performance in 11 problems, followed by IM-C-MOEA in 6 problems, RVEA and POCEA in 4 problems each, MOEA/DD in 3 problems, and LMOCSO in 2 problems.

IM-C-MOEA/D was found to be better in the unconstrained problem (RWMOP 9), with many objectives (RWMOP 11), and in the context of low inequality constraints in general (RWMOP 1-21, 25-29). IM-C-MOEA performed closest to its version with decomposition in terms of performance, excelling in finding feasible solutions to problems with equality constraints (RWMOP 23) and many constraints (RWMOP 35). POCEA performed better on problems with equality restrictions (RWMOP 1, 22-24). RVEA and MOEA/DD performed better on problems with more inequality restrictions and decision variables (RWMOP 30-35). 

The performance of RWMOPs 28, 33, and 34 can be attributed to the inherent features of these problem sets. RWMOP 28 involves the simultaneous management of inequality and equality constraints. Furthermore, RWMOP 33 and 34 involve 30 decision variables and 29 inequality constraints. Such scenarios present challenges in identifying feasible solutions, highlighting the need for effective constraint-handling techniques and optimization strategies to address such constrained optimization situations.

IM-MOEA/D was designed to deal with large-scale (up to 600 decision variables) and many objectives (up to 6) optimization problems. Adding a constraint handling technique, IM-C-MOEA/D can extend to real-world CMOPs.

\begin{table*}[htbp]
\scalefont{0.97}
\centering
\setlength{\tabcolsep}{3pt} 
\label{tab:selected}
\caption{Mean and standard deviation of the Hypervolume metric on RWMOP1-35. Best performances are Highlighted.}
\begin{tabular}{@{}ccccccc@{}}
\hline
Problem       & MOEADD                                              & RVEA                                                & LMOCSO                                        & POCEA                                         & IM-C-MOEA                                           & IM-C-MOEA/D                               \\ \hline
\multicolumn{7}{c}{Mechanical Design Problems}                                                                                                                                                                                                                                                                              \\
RWMOP1        & 9.5415e-2 (1.36e-2) $-$                             & 5.8494e-1 (6.41e-3) $\approx$                       & 1.6634e-1 (3.78e-1) $-$                       & \hl{5.9008e-1 (5.44e-3) +} & 4.6051e-1 (3.13e-2) $-$                             & 5.8560e-1 (7.39e-3)                       \\
RWMOP2        & 1.6658e-1 (1.38e-1) $-$                             & 1.2969e-1 (1.07e-1) $-$                             & 0.0000e+0 (0.00e+0) $-$                       & 1.0722e-1 (1.07e-1) $-$                       & \hl{3.6766e-1 (7.79e-3) +}       & 3.3757e-1 (3.56e-2)                       \\
RWMOP3        & 1.8434e-1 (1.68e-1) $-$                             & 1.3640e-1 (1.32e-1) $-$                             & 1.2892e-1 (1.63e-1) $-$                       & 2.7468e-1 (2.55e-1) $-$                       & \hl{8.6356e-1 (6.18e-3) +}       & 7.1513e-1 (4.62e-2)                       \\
RWMOP4        & 1.6516e-1 (1.68e-1) $-$                             & 1.2465e-1 (2.52e-1) $-$                             & 2.0180e-1 (2.77e-1) $-$                       & 7.8000e-2 (1.90e-1) $-$                       & \hl{7.8057e-1 (2.14e-2) +}       & 5.0765e-1 (9.09e-2)                       \\
RWMOP5        & 4.2741e-1 (1.37e-3) $\approx$                       & 4.1089e-1 (7.09e-3) $-$                             & 2.3936e-1 (6.63e-4) $-$                       & 4.1987e-1 (4.97e-3) $-$                       & 4.2608e-1 (7.55e-4) $-$                             & \hl{4.2795e-1 (8.56e-4)} \\
RWMOP6        & \hl{2.7637e-1 (4.00e-4) +}       & 2.7520e-1 (1.97e-4) +                             & 3.7843e-3 (2.07e-2) $-$                       & 2.7163e-1 (2.53e-3) $-$                       & 2.7533e-1 (2.31e-4) +                             & 2.7468e-1 (3.49e-4)                       \\
RWMOP7        & 4.7792e-1 (1.89e-3) $-$                             & 4.7329e-1 (2.52e-3) $-$                             & \hl{4.8097e-1 (1.87e-4) +} & 4.7703e-1 (2.58e-3) $-$                       & 4.7110e-1 (1.89e-3) $-$                             & 4.7906e-1 (2.08e-3)                       \\
RWMOP8        & 1.8588e-2 (1.98e-3) $-$                             & \hl{2.4442e-2 (6.91e-4) +}       & 2.1393e-2 (8.22e-4) $-$                       & 1.0487e-3 (4.00e-3) $-$                       & 2.3507e-2 (3.31e-4) $-$                             & 2.3871e-2 (4.29e-4)                       \\
RWMOP9        & 5.2956e-2 (4.20e-16) $-$                            & 4.0183e-1 (1.72e-3) $-$                             & 5.4659e-2 (1.11e-3) $-$                       & 3.6764e-1 (8.82e-3) $-$                       & 3.6739e-1 (8.50e-3) $-$                             & \hl{4.0321e-1 (2.05e-3)} \\
RWMOP10       & 7.8343e-2 (1.61e-14) $-$                            & 8.3741e-1 (3.01e-3) $-$                             & 1.3234e-1 (1.24e-2) $-$                       & 8.3582e-1 (3.81e-3) $-$                       & 8.3546e-1 (4.69e-3) $-$                             & \hl{8.4480e-1 (5.81e-4)} \\
RWMOP11       & 3.5929e-2 (9.13e-3) $-$                             & 6.3899e-2 (5.51e-3) $-$                             & 1.7825e-2 (3.39e-3) $-$                       & 6.5821e-2 (4.27e-3) $-$                       & 8.4717e-2 (2.04e-3) $-$                             & \hl{9.0599e-2 (1.47e-3)} \\
RWMOP12       & 8.6529e-2 (5.03e-2) $-$                             & 3.7746e-1 (1.61e-2) $-$                             & 1.9747e-1 (1.21e-1) $-$                       & 1.0165e-1 (1.32e-1) $-$                       & \hl{4.9200e-1 (1.73e-2) +}       & 4.0147e-1 (5.92e-2)                       \\
RWMOP13       & 7.1196e-2 (5.16e-3) $-$                             & 8.7233e-2 (1.94e-4) $-$                             & 3.6925e-2 (3.16e-2) $-$                       & 2.5287e-2 (3.66e-2) $-$                       & 8.6440e-2 (2.76e-4) $-$                             & \hl{8.7439e-2 (1.99e-4)} \\
RWMOP14       & 5.3019e-1 (2.89e-2) +                             & 5.9811e-2 (2.95e-2) $-$                             & 2.8898e-1 (1.54e-1) $-$                       & 6.4845e-2 (2.82e-2) $-$                       & \hl{5.3033e-1 (1.75e-2) +}       & 4.9617e-1 (3.39e-2)                       \\
RWMOP15       & 1.5010e-1 (1.02e-1) $-$                             & 4.9408e-1 (8.41e-2) +                             & 0.0000e+0 (0.00e+0) $-$                       & 4.8424e-1 (1.34e-1) +                       & \hl{4.9915e-1 (6.04e-3) +}       & 2.8940e-1 (7.85e-2)                       \\
RWMOP16       & 7.9114e-2 (6.47e-5) $-$                             & \hl{7.3781e-1 (5.49e-3) +}       & 2.3627e-1 (2.04e-2) $-$                       & 6.9917e-1 (1.51e-2) +                       & 7.1653e-1 (1.01e-2) +                             & 5.6697e-1 (7.51e-2)                       \\
RWMOP17       & 2.5019e-1 (1.27e-2) +                             & 1.8445e-1 (4.02e-2) $\approx$                       & 1.5358e-1 (1.08e-1) $\approx$                 & 0.0000e+0 (0.00e+0) $-$                       & 1.9777e-1 (2.98e-2) $\approx$                       & 1.9138e-1 (4.20e-2)                       \\
RWMOP18       & 3.7327e-2 (2.82e-4) $-$                             & 4.0146e-2 (9.09e-5)  $\approx$                            & 4.0171e-2 (7.79e-5) $\approx$ & 4.0252e-2 (8.40e-5) $\approx$                       & 3.8177e-2 (3.28e-4) $-$                             & \hl{4.0302e-2 (8.05e-5)}                       \\
RWMOP19       & 2.7191e-1 (2.07e-2) $-$                             & 2.5669e-1 (2.26e-2) $-$                             & 0.0000e+0 (0.00e+0) $-$                       & 1.1002e-3 (6.03e-3) $-$                       & 3.2244e-1 (6.73e-3) $\approx$ & \hl{3.2377e-1 (4.52e-3)}                       \\
RWMOP20       & 0.0000e+0 (0.00e+0) $\approx$                       & 0.0000e+0 (0.00e+0) $\approx$                       & 0.0000e+0 (0.00e+0) $\approx$                 & 0.0000e+0 (0.00e+0) $\approx$                 & 0.0000e+0 (0.00e+0) $\approx$                       & 0.0000e+0 (0.00e+0)                       \\
RWMOP21       & 2.9893e-2 (1.27e-3) $-$                             & 3.1280e-2 (3.05e-4) $\approx$ & 2.9334e-2 (4.27e-5) $-$                       & 3.1311e-2 (7.05e-5) $\approx$                 & 3.1130e-2 (2.39e-4) $-$                             & \hl{3.1405e-2 (6.63e-5)}                       \\
\multicolumn{7}{c}{Chemical Engineering Problems}                                                                                                                                                                                                                                                                           \\
RWMOP22       & 0.0000e+0 (0.00e+0) $\approx$                       & 0.0000e+0 (0.00e+0) $\approx$                       & 0.0000e+0 (0.00e+0) $\approx$                 & \hl{2.8788e-1 (4.32e-1) +} & 0.0000e+0 (0.00e+0) $\approx$                       & 0.0000e+0 (0.00e+0)                       \\
RWMOP23       & 1.6921e-1 (1.60e-1) +                             & 0.0000e+0 (0.00e+0) $\approx$                       & 0.0000e+0 (0.00e+0) $\approx$                 & \hl{1.0249e+0 (2.08e-1) +} & 1.1034e-1 (1.59e-1) +                             & 0.0000e+0 (0.00e+0)                       \\
RWMOP24       & 3.3333e-2 (1.83e-1) $\approx$                       & 0.0000e+0 (0.00e+0) $\approx$                       & 0.0000e+0 (0.00e+0) $\approx$                 & \hl{2.0518e+5 (7.65e+5) +} & 0.0000e+0 (0.00e+0) $\approx$                       & 0.0000e+0 (0.00e+0)                       \\
\multicolumn{7}{c}{Process, Design, and Synthesis Problems}                                                                                                                                                                                                                                                                  \\
RWMOP25       & 9.0909e-2 (6.38e-16) $-$                            & 2.3564e-1 (9.68e-4) $-$                             & 2.3450e-1 (1.05e-3) $-$                       & 2.3502e-1 (1.90e-3) $-$                       & 2.1984e-1 (2.41e-3) $-$                             & \hl{2.3916e-1 (1.36e-3)} \\
RWMOP26       & 1.4428e-1 (3.49e-2) $-$                             & 1.2908e-1 (7.21e-3) $-$                             & 3.0328e-3 (1.66e-2) $-$                       & 1.1034e-1 (6.18e-2) $-$                       & 1.6593e-1 (2.45e-2) $\approx$                       & \hl{1.7561e-1 (1.94e-2)} \\
RWMOP27       & 5.8817e+0 (1.30e+1) $-$                             & 4.3310e+4 (1.57e+4) $-$                             & \hl{1.9076e+6 (6.94e+6) +} & 1.9727e+4 (1.19e+4) $\approx$                 & 1.9524e+2 (2.44e+2) $-$                             & 1.4589e+5 (4.60e+5)                       \\
RWMOP28       & 0.0000e+0 (0.00e+0) $\approx$                       & 0.0000e+0 (0.00e+0) $\approx$                       & 0.0000e+0 (0.00e+0) $\approx$                 & 0.0000e+0 (0.00e+0) $\approx$                 & 0.0000e+0 (0.00e+0) $\approx$                       & 0.0000e+0 (0.00e+0)                       \\
RWMOP29       & 7.3212e-1 (8.56e-2) $\approx$ & 5.9826e-1 (1.40e-1) $-$                             & 0.0000e+0 (0.00e+0) $-$                       & 2.4476e-1 (2.21e-1) $-$                       & 6.9427e-1 (1.18e-1) $-$                             & \hl{7.5814e-1 (3.19e-2)}                       \\
\multicolumn{7}{c}{Power Electronics Problems}                                                                                                                                                                                                                                                                              \\
RWMOP30       & 1.7041e-1 (2.75e-1) $\approx$                       & \hl{3.3266e-1 (3.11e-1) +}       & 0.0000e+0 (0.00e+0) $-$                       & 0.0000e+0 (0.00e+0) $-$                       & 3.4240e-2 (1.07e-1) $\approx$                       & 1.0268e-1 (2.36e-1)                       \\
RWMOP31       & \hl{2.1301e-1 (2.87e-1) +}       & 1.6258e-1 (2.67e-1) +                             & 0.0000e+0 (0.00e+0) $\approx$                 & 0.0000e+0 (0.00e+0) $\approx$                 & 1.5029e-2 (5.78e-2) $\approx$                       & 2.0911e-2 (1.15e-1)                       \\
RWMOP32       & \hl{3.8724e-1 (3.79e-1) +}       & 3.3306e-1 (3.55e-1) +                             & 0.0000e+0 (0.00e+0) $-$                       & 4.4262e-2 (1.53e-1) $\approx$                 & 5.5420e-2 (1.53e-1) $\approx$                       & 9.9896e-2 (2.59e-1)                       \\
RWMOP33       & 0.0000e+0 (0.00e+0) $\approx$                       & 0.0000e+0 (0.00e+0) $\approx$                       & 0.0000e+0 (0.00e+0) $\approx$                 & 0.0000e+0 (0.00e+0) $\approx$                 & 0.0000e+0 (0.00e+0) $\approx$                       & 0.0000e+0 (0.00e+0)                       \\
RWMOP34       & 0.0000e+0 (0.00e+0) $\approx$                       & 0.0000e+0 (0.00e+0) $\approx$                       & 0.0000e+0 (0.00e+0) $\approx$                 & 0.0000e+0 (0.00e+0) $\approx$                 & 0.0000e+0 (0.00e+0) $\approx$                       & 0.0000e+0 (0.00e+0)                       \\
RWMOP35       & 3.2693e-2 (1.26e-1) $\approx$                       & \hl{4.8730e-2 (1.29e-1) +}                             & 0.0000e+0 (0.00e+0) $\approx$                 & 0.0000e+0 (0.00e+0) $\approx$                 & 2.7882e-2 (7.35e-2) +       & 0.0000e+0 (0.00e+0)                       \\ \hline
+/-/$\approx$ & 6/19/10                                             & 8/16/11                                             & 2/22/11                                       & 6/19/10                                        & 10/13/12                                            &                                           \\ \hline
\end{tabular}
\end{table*}

\section{Conclusion}
\label{sec:conclusion}

This paper proposed an inverse modeling CMOEA based on decomposition (IM-C-MOEA/D) for dealing with CMOPs with multiple or many objectives. This model uses a $k$-means scheme for clustering within the objective space, the selection criterion based on a global replacement to choose the most appropriate weight vector from the population, and a constraint handling approach to deal with CMOPs. IM-C-MOEA/D reached competitive performance compared to 6 state-of-the-art CMOEAs in most of the 35 real-world CMOPs \cite{kumar2021benchmark} with different inequality and equality constraints.

IM-C-MOEA/D is an improved version of IM-MOEA/D, inheriting favorable features such as mapping the objective space to the decision space. Despite the achievements, we can observe limitations in the IM-C-MOEA/D, which raise possibilities for future lines of research:
\begin{enumerate}
    \item Adapting the weight vectors might improve the diversity of the solutions over the evolutionary process since the shape of the Pareto front directly affects the performance of MOEAs based on decomposition;
    
    \item IM-C-MOEA/D has various parameters. Thus, it may be worth investigating the influence of parameter setting on the model performance;

    \item Exploring constraint-handling techniques to reach feasible solutions in problems that involve combinations of inequality and equality constraints and numerous decision variables.

\end{enumerate}



\bibliographystyle{IEEEtran}
\bibliography{references}

\end{document}